\documentclass[11pt,a4paper]{article}
\usepackage[hyperref]{acl2020}
\usepackage{times}
\usepackage{latexsym}
\usepackage{graphicx}

\usepackage{microtype}

\aclfinalcopy 

\title{Scaling Native Language Identification with Transformer Adapters}

\author{Ahmet Yavuz Uluslu \\
  Universität Zürich \& PRODAFT \\
  \texttt{ahmetyavuz.uluslu@uzh.ch} \\\And
  Gerold Schneider  \\
  Universität Zürich \\
  \texttt{gschneid@cl.uzh.ch} \\}

\date{}

\begin{document}
\maketitle
\begin{abstract}
Native language identification (NLI) is the task of automatically identifying the native language (L1) of an individual based on their language production in a learned language. It is useful for a variety of purposes including marketing, security and educational applications. NLI is usually framed as a multi-class classification task, where numerous designed features are combined to achieve state-of-the-art results. Recently deep generative approach based on transformer decoders (GPT-2) outperformed its counterparts and achieved the best results on the NLI benchmark datasets. We investigate this approach to determine the practical implications compared to traditional state-of-the-art NLI systems. We introduce transformer adapters to address memory limitations and improve training/inference speed to scale NLI applications for production.
\end{abstract}

\section{Introduction}
Native Language Identification (NLI) is the task of automatically identifying the native language (L1) of an individual based on their writing or speech in another language (L2). It is used for a variety of purposes including marketing, security and educational applications. The growing interest in NLI from various research fields can be partly attributed to the outstanding performance of automated NLI systems against human annotators. A study of human performance in NLI \citep{malmasi2015oracle} showed that NLI systems perform in the 80\%–90\% accuracy range while humans achieved 37.3\% average accuracy. The experimentation also constrained the number of considered languages and texts to enable human competition. 

NLI is most commonly framed as a multi-class classification problem. For text-based NLI, features are extracted from written resources produced by non-native speakers to train a classification model. The underlying hypothesis is that the L1 influences learners’ second language writing as a result of the language transfer effect \citep{odlin1989language}. A variety of feature types have been explored to capture distinct features of the language interference phenomenon: spelling errors \citep{koppel2005automatically, chen2017improving}; word and lemma n-grams \citep{tetreault2013report}; character n-grams \citep{kulmizev2017power}, dependency parsing and morphosyntax \citep{cimino2013linguistic}. As seen by the two shared tasks on the NLI task organized in 2013 and 2017, the combination of such features produces the best outcomes for NLI \citep{tetreault2013report, malmasi2017report}. The top ranked systems made use of Support Vector Machine (SVM) models trained on a diverse set of linguistic features that can capture word, sentence and document level characteristics. \citep{markov2017cic, cimino2017stacked}

Deep neural network based approaches were also considered for the NLI task. \citet{bjerva2017neural} experimented with deep residual networks (DNN), long short-term memory networks (LSTM) and continuous bag-of-words embeddings to create meta classifiers. \citet{li2017classifier} built an ensemble of single-feature SVMs fed into a multi-layer perceptron (MLP). \citet{habic2020multitask} integrated multi-task learning into convolutional neural networks (CNN) to create shared representations from multiple datasets. The studies concluded that traditional methods, i.e., SVM with engineered features, appear to work better than deep learning-based standalone and meta-classification approaches. Recently  \citet{lotfi2020deep} introduced the deep generative modelling approach to NLI which consists of fine-tuning a GPT-2 model to identify each language. Their method outperforms traditional machine learning approaches and currently achieves the best results on the benchmark NLI datasets.

\pagebreak

The contributions of the work presented here are the following: (i) We investigate the resource requirements and inference performance for the deep generative approach in comparison to traditional state-of-the-art NLI systems, and (ii) we introduce ProDAPT, transformer adapters based on deep generative model to optimize memory and storage space, and (iii) we evaluate our approach on the NLI task and explore the tradeoffs.

\section{Related Work}
\textbf{Deep Generative Approach.} 
OpenAI’s Generative Pre-trained Transformer-2 (GPT-2) is a unidirectional transformer-based language model pretrained on 40 GB of text data with the objective of predicting the next word given the context \citep{radford2019language}. It can generate coherent paragraphs of text and achieves state-of-the-art performance on many language modelling benchmarks without any task-specific training. Instead of training a classifier, the deep generative approach finetunes a generative model (GPT-2) on texts written by native speakers of each language (L1) to capture peculiarities of language transfer \citep{lotfi2020deep}. After training N (number of target languages) models to learn the characteristics of each L1, they can be used to discriminate between unseen text samples based on the language model (LM) loss. The least LM loss is expected from the model that is trained on the same class (L1). Although there are examples of the LM loss as a ranking feature for other tasks such as substitute selection for text simplification \citep{uluslu2022automatic}, the deep generative approach is the first method to use such value as the only discriminator for text classification.

\begin{figure}[h]
\includegraphics[width=0.5\textwidth]{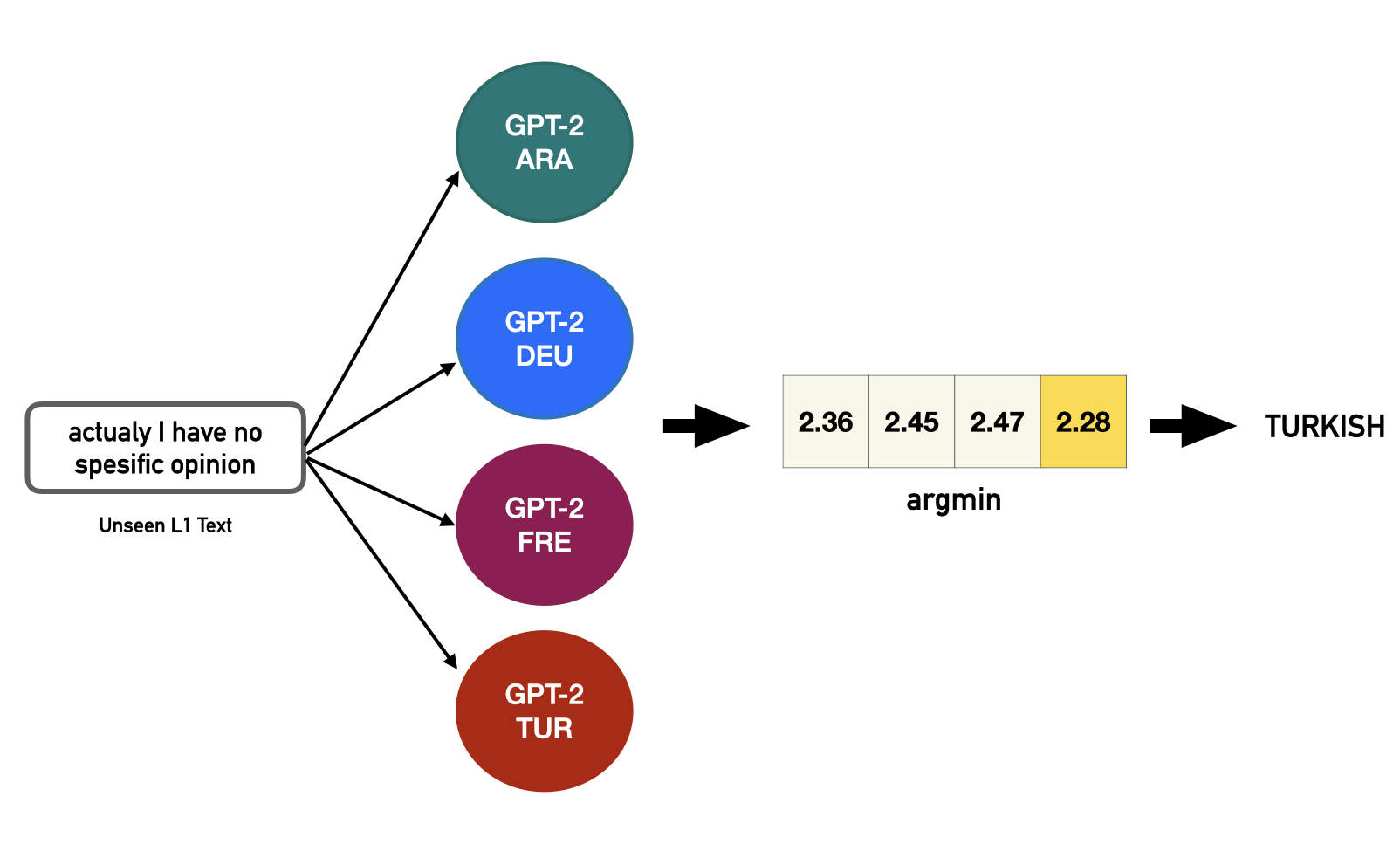}
\caption{An example inference of an unseen text written by a Turkish native speaker.}
\label{fig:figure2}
\end{figure}

\textbf{Transformer Adapters.} Adapters have been introduced as an alternative lightweight fine-tuning strategy that achieves equal performance to full
fine-tuning on most tasks \citep{houlsby2019parameter}. They consist of a small set of additional newly initialized weights at every layer of the transformer. While the rest of the pretrained parameters of the large model are kept frozen during the fine-tuning process, these new parameters are actively trained on the target task. Efficient parameter sharing between tasks is possible by training several task-specific and language-specific adapters for the same model, which can be exchanged and combined afterwards. The code base for different state-of-the-art adapter architectures was integrated into the transformers library and released under the name adapter-transformers \citep{pfeiffer2020adapterhub}. Recently, their adapters implementation started to support generative and seq2seq models such as GPT-2 \citep{sterz2021adapters}. 

\section{Data}
We evaluate our approach on the most commonly used dataset in NLI research:
the ETS Corpus of Non-Native Written English (TOEFL11) \citep{blanchard2013toefl11}. The dataset contains 1,100 essays in English written by native speakers (L1) of 11 different languages: Arabic (ARA), Chinese (CHI), French (FRE), German (GER), Hindi (HIN), Italian (ITA), Japanese (JPN), Korean (KOR), Spanish (SPA), Telugu (TEL), and Turkish (TUR). In total there are 12,100 essays with on average 348 tokens per essay. The essays were written in response to eight different writing prompts, all of which appear in all 11 L1 groups, by authors with low, medium, or high English proficiency. The dataset is considered a benchmark dataset for NLI and was used in two shared tasks on the NLI task \citep{tetreault2013report, malmasi2017report}.

\section{Methodology}
We investigate the resource requirements and inference speed for the deep generative approach in three different subsections: storage requirements, memory requirements and inference speed.

\textbf{Storage Requirements.} The full deep generative approach from \citet{lotfi2020deep} finetunes 11 gpt2-medium models to cover every language in the TOEFL11 dataset. A fine-tuned gpt2-medium model requires 1.4 GB of storage space. The training process with the early stopping of three validations is expected to take up to 61.6 GB of storage space upon completion. The inference (test) stage only requires the best-performing models for every language to be kept. Therefore, 15.4 GB of storage space in total is needed to fully store the system.  

\textbf{Memory Requirements.} To reach the final inference (target native language), each model needs to calculate the LM loss value for the given input. To fully load the parameters of 11 models in memory, 16.6 GB of GPU memory is required.

\textbf{Speed Constraints.} The deep generative approach does not inherently support parallelism. Further configuration in multi-threading and memory management is needed to accommodate different models in the GPU. The final inference still needs to be calculated in CPU time (argmin) and requires all model calculations to be completed beforehand. The number of models loaded simultaneously is constrained by the GPU requirements. In case all models are not available simultaneously, the memory operations to load/unload language models are part of the inference process and directly interfere with the performance. 

\textbf{Adapter Configurations.}
Two different configurations have been proposed for transformer adapters \citep{houlsby2019parameter, pfeifferconfig}. \citet{adapterdrop} investigated the efficiency of two adapter architectures at training and inference time and found that they achieved comparable performance. \citet{sterz2021adapters} compared the performance of two architectures for adapter-based GPT-2 models on the GLUE benchmark and reported on-par results on different tasks. We also experimented with both architectures and found that \citet{houlsby2019parameter} produced better results for standalone GPT-2 models trained on NLI data and decided to implement our full architecture based on this configuration. 

\begin{figure}[h]
\includegraphics[width=0.47\textwidth]{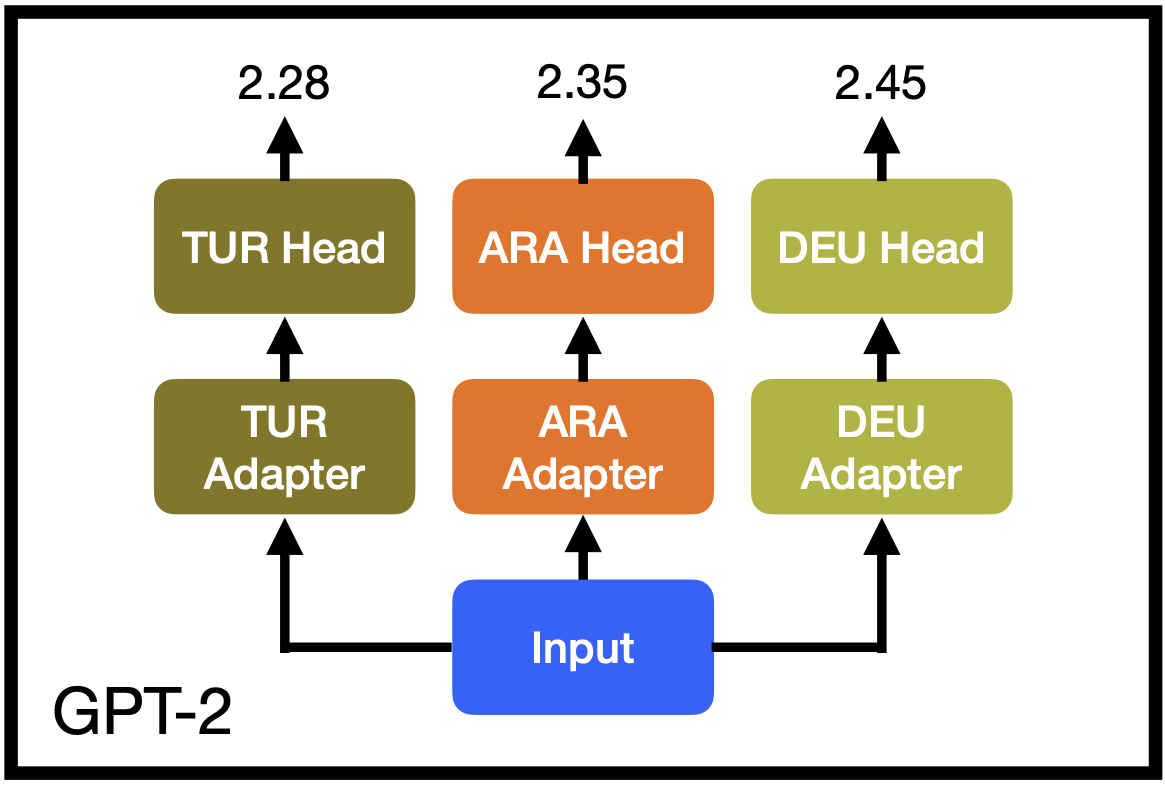}
\caption{The ProDAPT architecture}
\label{fig:figure2}
\end{figure}

The number of target languages for the NLI task can increase depending on the use case. In forensic linguistic investigations, particularly in the area of cybercrime, it may be necessary to cover up to 20-30 languages depending on the region and nature of crime, all of which may be in non-standardised forms, requiring the development of further models. To create a scalable NLI system, the memory bottleneck caused by the model size and the non-parallel nature of the deep generative approach should be addressed.

We implement ProDAPT architecture with transformer adapters to address these issues. We train an adapter for every L1 for 15 epochs with a learning rate of 1e-4.  The original pretrained weights for the GPT-2 are kept intact, and the L1 (target language) information is compressed into the newly initialized parameters and the classification head. The storage space required for every language model (adapter + head) decreases to 218.7 MB. All adapters and their classification heads are loaded into a single gpt2-medium model and share the pretrained weights. To support parallel inference, the input is replicated at the first layer with L1 adapters. For every adapter, calculations are completed in parallel until the classification head is reached. The GPU memory required to load the ProDAPT architecture is 4.1 GB and the storage space requirement is 2.4 GB. 

\section{Results and Discussion}
To compare the performance of our system with the state-of-the-art deep generative approach \citep{lotfi2020deep}, we report the results in terms of classification accuracy on the TOEFL11 test
set, as well as on the TOEFL11 dataset under 10-fold cross-validation (10FCV). We also report the storage and memory requirements to deploy our model in comparison to their approach. We use Titan T4 16 GB graphics card for the deep generative models and AMD EPYC 7702 64-Core CPU for Support Vector Machine (SVM) baselines where the GPU is not required. We think this is an acceptable choice since our work focuses on creating scalable NLI systems that can be easily deployed with widely available GPUs. We measure the inference speed with the time spent between CUDA events until the inference is finalized. We warm up the GPU before the test and repeat the experiment 100 times to enable robust results. Since the deep generative approach does not require feature engineering, we create a unigram SVM baseline to ensure a fair comparison.
We compare the performance against the default deep generative approach where one model at a time can be loaded into the GPU and the LM loss for the batch (4) is computed linearly. 

\begin{table}[h!]
\centering
\resizebox{\columnwidth}{!}{%
\begin{tabular}{|l|c|c|c|}
\hline
Model        & \multicolumn{1}{l|}{Storage Space} & GPU Memory & Inference Speed \\ \hline
Unigram SVM       & 79.6 MB                             & [CPU]     & 84x             \\ \hline
Lotfi et al. (2020) & 15.95 GB                           & 16.6 GB    & x               \\ \hline
ProDAPT      & 2.4 GB                             & 4.1 GB     & 13x             \\ \hline

\end{tabular}%
}
\caption{Results in terms of the storage space, GPU memory and inference speed.}
\label{tab:my-table}
\end{table}

\begin{table}[h!]
\centering
\begin{tabular}{|l|c|c|}
\hline
Model        & \begin{tabular}[c]{@{}c@{}}TOEFL11\\ (test set)\end{tabular} & \begin{tabular}[c]{@{}c@{}}TOEFL11\\ 10FCV\end{tabular} \\ \hline
Unigram SVM       & 75.8                                                         & 76.6                                                    \\ \hline
Lotfi et al. (2020) & 89.0                                                         & 86.8                                                    \\ \hline
ProDAPT      & 84.2                                                         & 82.4                                                    \\ \hline

\end{tabular}
\caption{Results in terms of classification accuracy (\%) on the TOEFL11 dataset.}
\label{tab:my-table}
\end{table}
The results presented in Table 1 demonstrate how the ProDAPT architecture optimises the state-of-the-art hardware requirements. The full parallelization support enables a considerable increase of 13x in the inference speed in comparison to the deep generative approach. As the number of supported languages (L1) increases, we predict that this performance gap will widen and become more significant. The unigram SVM baseline proves to be the most lightweight approach, and it provides the best results in terms of the inference speed. However, in order to achieve comparable performance to that of the deep generative approach, other approaches need to make use of ensemble learning and feature engineering. The extensive feature engineering in state-of-the-art systems, which can include hundreds of linguistic features, also comes at the expense of inference and training speed. The results shown in Table 2 indicate that our system outperforms the baseline and achieves an on-par performance with the deep generative approach. We found that the performance decline in our approach was caused by the difficulty of distinguishing between similar language pairs, such as Hindi-Telugu and Japanese-Korean, also noted by \citet{lotfi2020deep}. We did not observe convergence problems based on the LM loss for any of the L1 models, but we speculate that the restricted number of parameters in the adapters results in a reduced capacity to capture discriminative features that can distinguish between similar languages compared to full model fine-tuning.

\section{Conclusion}
We proposed an efficient and scalable NLI system based on the state-of-the-art deep generative approach. The method consists of training a transformers adapter for every L1 which can be attached simultaneously to a GPT-2 model for parallel inference. We showed that it is possible to optimize the hardware requirements and the inference speed at the cost of a slight decrease in the model performance. 

\section*{Acknowledgments}

This research received funding through the innovation cheque programme from the Swiss Innovation Agency. It is part of the research collaboration between the University of Zurich and PRODAFT under the project name 59977.1 INNO-ICT "Author profiling to automatize attribution during cybercrime investigations“.

\bibliography{acl2020}

\begin{thebibliography}{22}
\expandafter\ifx\csname natexlab\endcsname\relax\def\natexlab#1{#1}\fi

\bibitem[{Bjerva et~al.(2017)Bjerva, Grigonyt{\.e}, {\"O}stling, and
  Plank}]{bjerva2017neural}
Johannes Bjerva, Gintar{\.e} Grigonyt{\.e}, Robert {\"O}stling, and Barbara
  Plank. 2017.
\newblock Neural networks and spelling features for native language
  identification.
\newblock In \emph{Proceedings of the 12th Workshop on Innovative Use of NLP
  for Building Educational Applications}, pages 235--239.

\bibitem[{Blanchard et~al.(2013)Blanchard, Tetreault, Higgins, Cahill, and
  Chodorow}]{blanchard2013toefl11}
Daniel Blanchard, Joel Tetreault, Derrick Higgins, Aoife Cahill, and Martin
  Chodorow. 2013.
\newblock Toefl11: A corpus of non-native english.
\newblock \emph{ETS Research Report Series}, 2013(2):i--15.

\bibitem[{Chen et~al.(2017)Chen, Strapparava, and Nastase}]{chen2017improving}
Lingzhen Chen, Carlo Strapparava, and Vivi Nastase. 2017.
\newblock Improving native language identification by using spelling errors.
\newblock In \emph{Proceedings of the 55th Annual Meeting of the Association
  for Computational Linguistics (Volume 2: Short Papers)}, pages 542--546.

\bibitem[{Cimino and Dell’Orletta(2017)}]{cimino2017stacked}
Andrea Cimino and Felice Dell’Orletta. 2017.
\newblock Stacked sentence-document classifier approach for improving native
  language identification.
\newblock In \emph{Proceedings of the 12th Workshop on Innovative Use of NLP
  for Building Educational Applications}, pages 430--437.

\bibitem[{Cimino et~al.(2013)Cimino, Dell’Orletta, Venturi, and
  Montemagni}]{cimino2013linguistic}
Andrea Cimino, Felice Dell’Orletta, Giulia Venturi, and Simonetta Montemagni.
  2013.
\newblock Linguistic profiling based on general--purpose features and native
  language identification.
\newblock In \emph{Proceedings of the Eighth Workshop on Innovative Use of NLP
  for Building Educational Applications}, pages 207--215.

\bibitem[{Habic et~al.(2020)Habic, Semenov, and Pasiliao}]{habic2020multitask}
Vuk Habic, Alexander Semenov, and Eduardo~L Pasiliao. 2020.
\newblock Multitask deep learning for native language identification.
\newblock \emph{Knowledge-Based Systems}, 209:106440.

\bibitem[{Houlsby et~al.(2019)Houlsby, Giurgiu, Jastrzebski, Morrone,
  De~Laroussilhe, Gesmundo, Attariyan, and Gelly}]{houlsby2019parameter}
Neil Houlsby, Andrei Giurgiu, Stanislaw Jastrzebski, Bruna Morrone, Quentin
  De~Laroussilhe, Andrea Gesmundo, Mona Attariyan, and Sylvain Gelly. 2019.
\newblock Parameter-efficient transfer learning for {NLP}.
\newblock In \emph{International Conference on Machine Learning}, pages
  2790--2799. PMLR.

\bibitem[{Koppel et~al.(2005)Koppel, Schler, and
  Zigdon}]{koppel2005automatically}
Moshe Koppel, Jonathan Schler, and Kfir Zigdon. 2005.
\newblock Automatically determining an anonymous author’s native language.
\newblock In \emph{International Conference on Intelligence and Security
  Informatics}, pages 209--217. Springer.

\bibitem[{Kulmizev et~al.(2017)Kulmizev, Blankers, Bjerva, Nissim, van Noord,
  Plank, and Wieling}]{kulmizev2017power}
Artur Kulmizev, Bo~Blankers, Johannes Bjerva, Malvina Nissim, Gertjan van
  Noord, Barbara Plank, and Martijn Wieling. 2017.
\newblock The power of character n-grams in native language identification.
\newblock In \emph{Proceedings of the 12th workshop on innovative use of NLP
  for building educational applications}, pages 382--389.

\bibitem[{Li and Zou(2017)}]{li2017classifier}
Wen Li and Liang Zou. 2017.
\newblock Classifier stacking for native language identification.
\newblock In \emph{Proceedings of the 12th Workshop on Innovative Use of NLP
  for Building Educational Applications}, pages 390--397.

\bibitem[{Lotfi et~al.(2020)Lotfi, Markov, and Daelemans}]{lotfi2020deep}
Ehsan Lotfi, Ilia Markov, and Walter Daelemans. 2020.
\newblock A deep generative approach to native language identification.
\newblock In \emph{Proceedings of the 28th International Conference on
  Computational Linguistics}, pages 1778--1783.

\bibitem[{Malmasi et~al.(2017)Malmasi, Evanini, Cahill, Tetreault, Pugh,
  Hamill, Napolitano, and Qian}]{malmasi2017report}
Shervin Malmasi, Keelan Evanini, Aoife Cahill, Joel Tetreault, Robert Pugh,
  Christopher Hamill, Diane Napolitano, and Yao Qian. 2017.
\newblock A report on the 2017 native language identification shared task.
\newblock In \emph{Proceedings of the 12th Workshop on Innovative Use of NLP
  for Building Educational Applications}, pages 62--75.

\bibitem[{Malmasi et~al.(2015)Malmasi, Tetreault, and Dras}]{malmasi2015oracle}
Shervin Malmasi, Joel Tetreault, and Mark Dras. 2015.
\newblock Oracle and human baselines for native language identification.
\newblock In \emph{Proceedings of the tenth workshop on innovative use of NLP
  for building educational applications}, pages 172--178.

\bibitem[{Markov et~al.(2017)Markov, Chen, Strapparava, and
  Sidorov}]{markov2017cic}
Ilia Markov, Lingzhen Chen, Carlo Strapparava, and Grigori Sidorov. 2017.
\newblock Cic-fbk approach to native language identification.
\newblock In \emph{Proceedings of the 12th Workshop on Innovative Use of NLP
  for Building Educational Applications}, pages 374--381.

\bibitem[{Odlin(1989)}]{odlin1989language}
Terence Odlin. 1989.
\newblock \emph{Language transfer}, volume~27.
\newblock Cambridge University Press Cambridge.

\bibitem[{Pfeiffer et~al.(2020{\natexlab{a}})Pfeiffer, R{\"u}ckl{\'e}, Poth,
  Kamath, Vuli{\'c}, Ruder, Cho, and Gurevych}]{pfeiffer2020adapterhub}
Jonas Pfeiffer, Andreas R{\"u}ckl{\'e}, Clifton Poth, Aishwarya Kamath, Ivan
  Vuli{\'c}, Sebastian Ruder, Kyunghyun Cho, and Iryna Gurevych.
  2020{\natexlab{a}}.
\newblock Adapterhub: A framework for adapting transformers.
\newblock \emph{arXiv preprint arXiv:2007.07779}.

\bibitem[{Pfeiffer et~al.(2020{\natexlab{b}})Pfeiffer, Vulić, Gurevych, and
  Ruder}]{pfeifferconfig}
Jonas Pfeiffer, Ivan Vulić, Iryna Gurevych, and Sebastian Ruder.
  2020{\natexlab{b}}.
\newblock \href {https://doi.org/10.48550/ARXIV.2005.00052} {Mad-x: An
  adapter-based framework for multi-task cross-lingual transfer}.

\bibitem[{Radford et~al.(2019)Radford, Wu, Child, Luan, Amodei, Sutskever
  et~al.}]{radford2019language}
Alec Radford, Jeffrey Wu, Rewon Child, David Luan, Dario Amodei, Ilya
  Sutskever, et~al. 2019.
\newblock Language models are unsupervised multitask learners.
\newblock \emph{OpenAI blog}, 1(8):9.

\bibitem[{Rücklé et~al.(2020)Rücklé, Geigle, Glockner, Beck, Pfeiffer,
  Reimers, and Gurevych}]{adapterdrop}
Andreas Rücklé, Gregor Geigle, Max Glockner, Tilman Beck, Jonas Pfeiffer,
  Nils Reimers, and Iryna Gurevych. 2020.
\newblock \href {https://doi.org/10.48550/ARXIV.2010.11918} {Adapterdrop: On
  the efficiency of adapters in transformers}.

\bibitem[{Sterz et~al.(2021)Sterz, Poth, R{\"u}ckl{\'e}, and
  Pfeiffer}]{sterz2021adapters}
Hannah Sterz, Clifton Poth, Andreas R{\"u}ckl{\'e}, and Jonas Pfeiffer. 2021.
\newblock Adapters for generative and seq2seq models in nlp.
\newblock \emph{Blog Post}.

\bibitem[{Tetreault et~al.(2013)Tetreault, Blanchard, and
  Cahill}]{tetreault2013report}
Joel Tetreault, Daniel Blanchard, and Aoife Cahill. 2013.
\newblock A report on the first native language identification shared task.
\newblock In \emph{Proceedings of the eighth workshop on innovative use of NLP
  for building educational applications}, pages 48--57.

\bibitem[{Uluslu(2022)}]{uluslu2022automatic}
Ahmet~Yavuz Uluslu. 2022.
\newblock Automatic lexical simplification for {Turkish}.
\newblock \emph{arXiv preprint arXiv:2201.05878}.

\end{thebibliography}
\bibliographystyle{acl_natbib}

\end{document}